\documentclass[preprint,11pt]{elsarticle}

\usepackage{amsmath}
\usepackage{amsfonts}
\usepackage{amssymb}
\usepackage{graphicx}
\usepackage{subfigure}
\usepackage{url}
\usepackage{lineno}

\begin{document}

\begin{frontmatter}

\title{\huge Image classification by visual bag-of-words \\refinement and reduction}

\author[ruc]{Zhiwu Lu \corref{cor}}
\ead{zhiwu.lu@gmail.com}
\author[pku]{Liwei Wang}
\ead{wanglw@cis.pku.edu.cn}
\author[ruc]{Ji-Rong Wen}
\ead{jirong.wen@gmail.com}
\address[ruc]{School of Information, Renmin University of China, Beijing 100872, China}
\address[pku]{Key Laboratory of Machine Perception (MOE), School of EECS, Peking University, Beijing 100871, China}
\cortext[cor]{Corresponding author. Tel: 86-10-62514562; Fax: 86-10-62514562.}

\begin{abstract}
This paper presents a new framework for visual bag-of-words (BOW) refinement and reduction to overcome the drawbacks associated with the visual BOW model which has been widely used for image classification. Although very influential in the literature, the traditional visual BOW model has two distinct drawbacks. Firstly, for efficiency purposes, the visual vocabulary is commonly constructed by directly clustering the low-level visual feature vectors extracted from local keypoints, without considering the high-level semantics of images. That is, the visual BOW model still suffers from the semantic gap, and thus may lead to significant performance degradation in more challenging tasks (e.g. social image classification). Secondly, typically thousands of visual words are generated to obtain better performance on a relatively large image dataset. Due to such large vocabulary size, the subsequent image classification may take sheer amount of time. To overcome the first drawback, we develop a graph-based method for visual BOW refinement by exploiting the tags (easy to access although noisy) of social images. More notably, for efficient image classification, we further reduce the refined visual BOW model to a much smaller size through semantic spectral clustering. Extensive experimental results show the promising performance of the proposed framework for visual BOW refinement and reduction.
\end{abstract}


\end{frontmatter}


\section{Introduction}

Inspired by the success of bag-of-words (BOW) in text information
retrieval, we can similarly represent an image as a histogram of
visual words through quantizing the local keypoints within the image into visual words, which is known as visual BOW in the areas of image analysis and computer vision. As an intermediate representation, the visual BOW model can help to reduce the semantic gap between the low-level visual features and the high-level semantics of images to some extent. Hence, many efforts have been made to apply the visual BOW model to image classification. In fact, the visual BOW model has been shown to give rise to encouraging results in image classification \cite{LSP06,MNJ08,LWL11,CIB13,lu2010combining,lu2011spatial,lu2011contextual,wang2009image}. In the following, we refer to the visual words as mid-level features to distinguish them from the low-level visual features and high-level semantics of images.

However, as reported in previous work
\cite{MJJ10,JXY09,ZZ07,BZM06,FP05}, the traditional visual BOW model has two distinct drawbacks. Firstly, for efficiency
purposes, the visual vocabulary is commonly constructed by directly
clustering \cite{lu2009generalized,lu2008comparing,lu2006iterative} the low-level visual feature vectors extracted from local
keypoints within images, without considering the
high-level semantics of images. That is, although the visual BOW
model can help to reduce the semantic gap to some extent, it still
suffers from the problem of semantic gap and thus may lead to
significant performance degradation in more challenging tasks (e.g., social image classification with larger intra-class variations).
Secondly, typically thousands of mid-level features are generated to
obtain better performance on a relatively large image dataset. Due
to such large vocabulary size, the subsequent image classification
may take sheer amount of time. This means that visual BOW reduction
becomes crucial for the efficient use of the visual BOW model in social image classification. In this paper, our main motivation is to simultaneously overcome these two drawbacks by proposing a new framework for visual BOW refinement and reduction, which will be elaborated in the following. It should be noted that these two drawbacks are usually considered separately in the literature \cite{JXY09,ZZ07,BZM06,FP05,MJJ10,GVS10,LYS09}. More thorough reviews of previous methods can be found in Section~\ref{sect:rw}.

To overcome the first drawback, we develop a graph-based method to
exploit the tags of images for visual BOW refinement. The basic idea
is to formulate visual BOW refinement as a multi-class
semi-supervised learning (SSL) problem. That is, we can regard each visual word as a ``class" and thus take the visual BOW representation as the initial configuration of SSL. In this paper, we focus on solving this problem by the graph-based SSL techniques \cite{ZGL03,ZBLW04}. Considering that graph construction is the
key step of graph-based SSL, we construct a new $L_1$-graph over
images with structured sparse representation by exploiting both the
original visual BOW model and the tags of images, which is different
from the traditional $L_1$-graph \cite{YW09,CYS10,CLY09} constructed
only with sparse representation \cite{SED05,WYG09}. Through
semi-supervised learning with such new $L_1$-graph, we can
explicitly utilize the tags of images (i.e. high-level semantics) to
reduce the semantic gap associated with the visual BOW model to some
extent.

Although the semantic information can be exploited for visual BOW
refinement using the above graph-based SSL, the vocabulary size of the refined visual BOW model remains unchanged. Hence, given a large initial visual vocabulary, the subsequent image classification may still take sheer amount of time. For efficient image classification, we further reduce the refined visual BOW model to a much smaller size through spectral clustering \cite{NJW02,WYH06,LI10,lu2013exhaustive} over mid-level features. A reduced set of high-level features is generated by regarding each cluster of mid-level features as a new higher level feature. Moreover, since the tags of images has been incorporated into the refined visual BOW model, we indirectly consider the semantic information in visual BOW reduction by using the refined visual BOW model for spectral clustering. In the following, our method is thus called as semantic spectral clustering. When tested in image classification, the reduced set of high-level features is shown to cause much less time but with little performance degradation.

\begin{figure*}[t]
\vspace{0.00in}
\begin{center}
\includegraphics[width=0.98\textwidth]{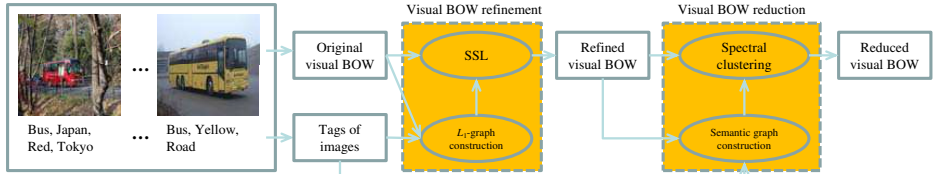}
\end{center}
\vspace{-0.25in} \caption{Illustration of the proposed framework for
visual BOW refinement and reduction by exploiting the tags of
images. The visual BOW refinement is newly proposed in the present
work (i.e. our main contribution), while the visual BOW reduction is
mainly proposed in our short conference version \cite{LP11}.} \label{Fig.1}
\vspace{-0.00in}
\end{figure*}

In summary, we propose a new framework for visual BOW refinement and reduction, and the system overview is illustrated in Figure~\ref{Fig.1}. In fact, upon our short conference version \cite{LP11}, we have made two extra contributions (also see Figure~\ref{Fig.1}): visual BOW refinement, and semantic graph construction for visual BOW reduction. Moreover, the advantages of the proposed framework can be summarized as follows: (1) our visual BOW refinement and reduction are both efficient even for large image datasets; (2) when the global visual features are fused for image classification, we can \emph{obtain the best results so far} (to the best of our knowledge) on the PASCAL VOC'07 \cite{EVW07} and MIR FLICKR \cite{HL08} benchmark datasets, as shown in our later experiments; (3) although only tested in image classification, our visual BOW refinement and reduction can be extended to other tasks (e.g. image annotation).

The remainder of this paper is organized as follows.
Section~\ref{sect:rw} provides an overview of related work. In
Section~\ref{sect:refine}, we develop a graph-based method to
explicitly utilize the tags of images for visual BOW refinement. In
Section~\ref{sect:reduct}, the refined visual BOW model is further
reduced to a much smaller size through semantic spectral clustering.
In Section~\ref{sect:exp}, the refined and reduced visual BOW models
are evaluated by directly applying them to image classification.
Finally, Section~\ref{sect:con} gives the conclusions.

\section{Related work}
\label{sect:rw}

Since visual BOW refinement and reduction, and structured
sparse representation are considered in the proposed framework,
we will give an overview of these techniques in the following.

\subsection{Visual BOW refinement}
\label{sect:rw:ref}

In this paper, visual BOW refinement refers to adding the semantics
of images to the visual BOW model. The main goal of visual BOW refinement is to bridge the semantic gap associated with the traditional visual BOW model. To the best of our knowledge, there exist at least two types of semantics which can be exploited for visual BOW refinement: (1) the constraints with respect to local keypoints, and (2) the tags of images. Derived from prior knowledge (e.g. the wheel and window of a car should occur together), the constraints with respect to local keypoints can be directly used as the clustering conditions for clustering-based visual BOW generation \cite{MJJ10}. However, the main disadvantage of this approach is that the constraints are commonly very expensive to obtain in practice. In contrast, the tags of images are much easier to access for social image collections. Hence, in this paper, we focus on exploiting the tags of images for visual BOW refinement.

Unlike our idea of utilizing the tags of images to refine the visual
BOW model and then improve the performance of image classification,
this semantic information can also be directly used as features for image classification. For example, by combining the tags of images with the global (e.g. color histogram) and local (e.g. BOW) visual
features, one influential work \cite{GVS10} has reported the best
classification results so far (to the best of our knowledge) on the PASCAL VOC'07 \cite{EVW07} and MIR FLICKR \cite{HL08} benchmark datasets. However, when the global visual features (actually much weaker than those used in \cite{GVS10}) are also considered for image classification in this paper, our later experimental results demonstrate that our method performs better than \cite{GVS10} on these two benchmark datasets.

Besides the tags of images, other types of information can also be used to bridge the semantic gap associated with the visual representation. In \cite{LWWZ08,LI10,HWP08,LSP06}, local or global spatial information is incorporated into the visual representation, which leads to obvious performance improvements in image classification. In \cite{XJS13,KAA13}, extra depth information is considered for image classification in the ImageCLEF 2013 Robot Vision Task. In \cite{KJK13}, inspired from the biological/cognitive models, a hierarchical structure is learnt for computer vision tasks. Although the goal of these approaches is the same as that of our method, we focus on utilizing the tags of images to bridge the semantic gap in this paper. In fact, the spatial, depth, or hierarchical information can be similarly added to our refined visual BOW model. For example, our refined visual BOW model can be used just as the original one to define spatial pyramid matching kernel \cite{LSP06}.

\subsection{Visual BOW reduction}
\label{sect:rw:red}

The goal of visual BOW reduction is to reduce the visual BOW model
of large vocabulary size to a much smaller size. This is mainly
motivated by the fact that a large visual BOW model causes sheer amount of time in image classification although it can achieve better
performance on a relatively large image dataset. In this paper, to handle this problem, we propose a semantic spectral clustering method for visual BOW reduction. The distinct advantage of our method is that the manifold structure of mid-level features can be preserved explicitly, unlike the traditional topic models  \cite{ZZ07,BZM06,FP05} for visual BOW reduction without considering such intrinsic
geometric information.  This is also the reason why our method significantly outperforms latent Dirichlet allocation \cite{BNJ03} (one of the most outstanding methods for visual BOW reduction in the literature) as shown in our later experiments.

Similar to our semantic spectral clustering, the diffusion map
method proposed in \cite{LYS09} can also exploit the intrinsic
geometric information for visual BOW reduction. However, the main disadvantage of this method is that it requires fine parameter tuning for graph construction which can significantly affect the performance of visual BOW reduction. In contrast, we construct semantic graphs in a parameter-free manner in this paper. As shown in our later experiments, our spectral clustering with semantic graphs can help to discover more intrinsic manifold structure of mid-level features and thus lead to obvious performance improvements. Moreover, since we focus on parameter-free graph construction for spectral clustering in this paper, we only adopt the commonly used technique introduced in \cite{NJW02}, without considering other spectral clustering techniques \cite{WYH06,LYS09,LI10,lu2013exhaustive} developed in the literature.

Although our visual BOW reduction can be regarded as kind of dimension
reduction over mid-level features, it has two distinct advantages over the traditional dimension reduction approaches \cite{LL06,YXZ07}
directly using spectral embedding. Firstly, each higher level feature learnt by our visual BOW reduction is actually a group of mid-level features which tend to be semantically related as shown in Figure~\ref{Fig.4}. However, the traditional dimension reduction approaches performing spectral embedding over all the data fail to give explicit explanation of each reduced feature, since they directly utilize the eigenvectors of the Laplacian matrix to form the new feature representation. Secondly, our visual BOW reduction by semantic spectral clustering over mid-level features takes much less time than the traditional dimension reduction approaches by spectral embedding with graphs over all the data.

\subsection{Structured sparse representation}

In this paper, we formulate visual BOW refinement as a multi-class
graph-based SSL problem. Considering that graph construction is the
key step of graph-based SSL, we develop a new $L_1$-graph
construction method using structured sparse representation, other
than the traditional $L_1$-graph construction method only using
sparse representation. As compared with sparse representation, our
structured sparse representation has a distinct advantage, i.e., the
extra structured sparsity can be induced into $L_1$-graph
construction and thus the noise in the data can be suppressed to the
most extent. In fact, the structured sparsity penalty used in this
paper is defined as $L_1$-norm Laplacian regularization, which is
formulated directly over all the eigenvectors of the normalized
Laplacian matrix. Hence, our new $L_1$-norm Laplacian regularization
is different from the $p$-Laplacian regularization \cite{ZS05} as an
ordinary $L_1$-generalization (with $p=1$) of the traditional
Laplacian regularization (see further comparison in Section
\ref{sect:refine:ssr}). In this paper, we focus on exploiting the
manifold structure of the data for $L_1$-graph construction with
structured sparse representation, regardless of other types of
structured sparsity \cite{EV11,JMO11} used in the literature.

\section{Visual BOW refinement}
\label{sect:refine}

This section presents our visual BOW refinement method in detail. We
first give our problem formulation for visual BOW refinement from a
graph-based SSL viewpoint, and then construct a new $L_1$-graph
using structured sparse representation for such graph-based SSL.
Finally, we provide the complete algorithm for our visual BOW
refinement based on the new $L_1$-graph.

\subsection{Problem formulation}

\begin{figure}[t]
\vspace{0.00in}
\begin{center}
\includegraphics[width=0.75\textwidth]{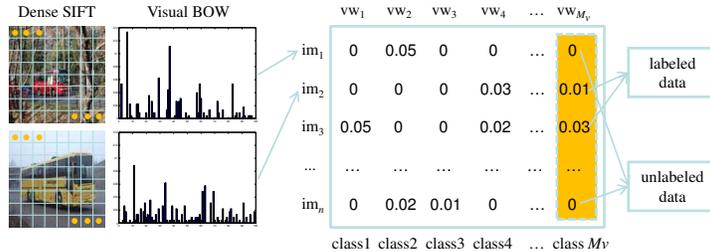}
\end{center}
\vspace{-0.25in} \caption{Illustration of visual BOW refinement from
a multi-class graph-based SSL viewpoint. Each column denote a visual
word $\mathrm{vw}_j (j=1,...,M_v)$, while each row denotes an image
$\mathrm{im}_i (i=1,...,n)$. Here, each column actually provides an initial configuration (e.g. labeled and unlabeled data along $\mathrm{vw}_{M_v}$) of graph-based SSL for a single class.} \label{Fig.2} \vspace{-0.0in}
\end{figure}

In this paper, we focus on visual BOW refinement by exploiting the tags of images, which are easy to access for social image collections. Similar to the formation of the visual BOW model, we generate a new textual BOW model with the tags of images. This means that our goal is actually to refine the visual BOW model based on the textual BOW model. As stated in Section I, we can transform visual BOW refinement into a multi-class SSL problem, which is also illustrated in Figure~\ref{Fig.2}. Although this multi-class problem can be solved by many other machine learning techniques, we only consider the graph-based SSL method \cite{ZBLW04} in this paper. The problem formulation is elaborated as follows.

Let $Y \in R^{n\times M_v}$ denote the visual BOW model and $A \in
R^{n\times n}$ denote the kernel (affinity) matrix computed over the
textual BOW model, where $n$ is the number of images and $M_v$ is
the number of visual words. In this paper, we only adopt linear
kernel to define the similarity matrix over the textual BOW model.
By directly setting the weight matrix $W=A$, we construct an
undirected graph $\mathcal{G}=\{\mathcal{V},W\}$ with its vertex set
$\mathcal{V}$ being the set of images. The normalized Laplacian
matrix of $\mathcal{G}$ is given by
\begin{eqnarray}
L=I-D^{-1/2}WD^{-1/2}, \label{eq:lap}
\end{eqnarray}
where $I$ is an identity matrix and $D$ is a diagonal matrix with
its $i$-th diagonal element being the sum of the $i$-th row of $W$.
The normalized Laplacian matrix $L$ is nonnegative definite.

Based on the above preliminary notations, the problem of visual BOW
refinement can be formulated from a multi-class graph-based SSL
viewpoint as illustrated in Figure~\ref{Fig.2}:
\begin{eqnarray}
F^*=\arg\min_F \frac{1}{2}||F-Y||_F^2 + \frac{\lambda}{2}
\mathrm{tr}(F^TLF), \label{eq:bowsr}
\end{eqnarray}
where $F \in R^{n\times M_v}$ denotes the refined visual BOW model,
$\lambda$ denotes the positive regularization parameter, and
$\mathrm{tr}(\cdot)$ denotes the trace of a matrix. The first term
of the above objective function is the fitting constraint, which
means a good $F$ should not change too much from the initial $Y$.
The second term is the smoothness constraint, which means that a
good $F$ should not change too much between similar images.
According to \cite{ZBLW04}, the above graph-based SSL problem has an
analytical solution:
\begin{eqnarray}
F^*=(I+\lambda L)^{-1}Y. \label{eq:bowsrsol}
\end{eqnarray}
We can clearly observe that \emph{the semantic information has been
added to the visual BOW model} by defining the normalized Laplacian
matrix $L$ using the tags of images (i.e. the textual BOW model).
More notably, such visual BOW refinement can effectively bridge the
semantic gap associated with the traditional visual BOW model, as
shown in our later experiments. The only disadvantage of the above
analytical solution is that it is not efficient for large image
datasets, since it has a time complexity of $O(n^3)$.

To apply our visual BOW refinement to large datasets, we have
to concern two key subproblems: how to construct the graph
efficiently, and how to solve the problem in Equation~(\ref{eq:bowsr}) efficiently. Moreover, since noisy tags may be used for our visual BOW refinement, we also need to ensure that our graph construction is noise-robust. These two subproblems will be addressed in the next two subsections, respectively.

\subsection{$L_1$-Graph construction with structured sparse representation}
\label{sect:refine:ssr}

Considering the important role of $\mathcal{G}$ in the above visual
BOW refinement, we first focus on graph construction over images. In
the literature, the $k$-nearest neighbor ($k$-NN) graph has been
widely used for graph-based SSL, since the problem in Equation~(\ref{eq:bowsr}) can be solved by the algorithm proposed in
\cite{ZBLW04} with linear time complexity based on this simple graph
($k \ll n$). However, the $k$-NN graph suffers from inherent
limitations (e.g. sensitivity to noise). To deal with the noise
(e.g. noisy tags here), a new $L_1$-graph construction method has
recently been developed based on sparse representation. The basic
idea of $L_1$-graph construction is to seek a sparse linear
reconstruction of each image with the other images. Unfortunately,
such $L_1$-graph construction may become infeasible since it takes sheer amount of time given a large data size $n$. To make a tradeoff,
we only consider the $k$ nearest neighbors of each image for sparse
linear reconstruction of this image, which thus becomes a much
smaller scale optimization problem ($k \ll n$). Moreover, to induce
the structured sparsity into $L_1$-graph construction and further
suppress the noise in the data, we exploit the manifold structure of
the $k$ nearest neighbors for sparse linear reconstruction of each
image. This $L_1$-graph construction will be elaborated as follows.

We start with the problem formulation for sparse linear
reconstruction of each image in its $k$-nearest neighborhood. Given
an image $x_i \in R^{M_t}$ ($i=1,...,n$) represented by the textual
BOW model (of the vocabulary size $M_t$), we suppose it can be
reconstructed using its $k$-nearest neighbors (their indices are
collected into $\mathcal{N}_k(i)$) as follows: $x_i = B_i \alpha_i +
\eta_i$, where $\alpha_i \in R^k$ is a vector that stores unknown
reconstruction coefficients, $B_i = [x_{i_1},x_{i_2},...,
x_{i_k}]_{i_j\in \mathcal{N}_k(i), j=1,...,k} \in R^{M_t \times k}$
is a dictionary with each column being regarded as a base, and
$\eta_i \in R^{M_t}$ is the noise term. Here, it should be noted
that we just follow the idea of \cite{WYG09} to introduce the noise
term $\eta_i$ into the linear reconstruction of $x_i$. The main
concern is that $x_i = B_i \alpha_i$ may not be exactly satisfied
since we have $M_t > k$ in this paper. Let $B'_i = [B_i, I] \in R^{M_t \times (M_t+k)}$
and $\alpha'_i = [\alpha^T_i, \eta^T_i ]^T$. The linear
reconstruction of $x_i$ can be transformed into: $ x_i = B'_i
\alpha'_i$, which is now an underdetermined system with respect to
$\alpha'_i$ since we always have $M_t < (M_t+k)$. According to
\cite{Donoho04}, if the solution (i.e. $\alpha'_i$) with respect to
$x_i$ is sparse enough, it can be recovered by solving the following
$L_1$-norm optimization problem:
\begin{eqnarray}
\min_{\alpha'_i}~~||\alpha'_i||_1,~~\mathrm{s.t.}~~x_i = B'_i
\alpha'_i, \label{eq:srori}
\end{eqnarray}
where $||\alpha'_i||_1$ is the $L_1$-norm of $\alpha'_i$. Given the
kernel (affinity) matrix $A=\{a_{ij}\}_{n\times n}$ computed over
the textual BOW model (also used as the weight matrix in Equation
(\ref{eq:lap})), we utilize the kernel trick and transform the
linear reconstruction of $x_i$ into: $(B_i^T x_i) = (B_i^T B_i)
\alpha_i + (B_i^T \eta_i)$. Let $y_i=B_i^T x_i=[a_{ji}]_{j\in
\mathcal{N}_k(i)} \in R^k$, $C_i=B_i^T B_i=[a_{jj'}]_{j,j'\in
\mathcal{N}_k(i)} \in R^{k\times k}$, and $\zeta_i=B_i^T \eta_i \in
R^k$. The original $L_1$-norm optimization problem in Equation~(\ref{eq:srori}) can be reformulated as:
\begin{eqnarray}
\min_{\alpha_i,\zeta_i}~~||[\alpha_i^T,\zeta_i^T]||_1,~~\mathrm{s.t.}~~y_i
= C_i \alpha_i +\zeta_i. \label{eq:sr}
\end{eqnarray}
Let $C'_i = [C_i, I] \in R^{k \times 2k}$ and $\alpha'_i =
[\alpha^T_i, \zeta^T_i ]^T$, the above problem can be further
transformed into:
\begin{eqnarray}
\min_{\alpha'_i}~~||\alpha'_i||_1,~~\mathrm{s.t.}~~y_i = C'_i
\alpha'_i, \label{eq:srsol}
\end{eqnarray}
which is similar to the original problem in Equation~(\ref{eq:srori}). This is a standard $L_1$-norm optimization problem, and we can solve it just as \cite{YW09,WYG09}. In this paper, we directly use the Matlab toolbox $\ell_{1}$-MAGIC\footnote{\url{http://users.ece.gatech.edu/~justin/l1magic/}}.

After we have obtained the reconstruction coefficients for all the
images by the above sparse linear reconstruction, the weight matrix
$W=\{w_{ij}\}_{n\times n}$ can be defined the same as \cite{YW09}:
\begin{eqnarray}
w_{ij} =\begin{cases}
|\alpha'_i(j')|,   & j\in\mathcal{N}_k(i),j'=\mathrm{index}(j,\mathcal{N}_k(i)); \\
0,   & \mathrm{otherwise}, \\
\end{cases}  \label{eq:l1wt}
\end{eqnarray}
where $\alpha'_i(j')$ denotes the $j'$-th element of the vector
$\alpha'_i$, and $j'=\mathrm{index}(j,\mathcal{N}_k(i))$ means that
$j$ is the $j'$-th element of the set $\mathcal{N}_k(i)$. By setting
the weight matrix $W=(W+W^T)/2$, we then construct an undirected
graph $\mathcal{G} = \{\mathcal{V}, W\}$ with the vertex set
$\mathcal{V}$ being the set of images. In the following, this graph
is called as $L_1$-graph, since it is constructed by
$L_1$-optimization.

In this $L_1$-graph, the similarity between images is defined as the
reconstruction coefficients of the sparse linear reconstruction
solution. However, the structured sparsity of these reconstruction
coefficients is actually ignored in such sparse representation. To
address this problem, we further induce the structured sparsity into
$L_1$-graph construction. In this paper, we only consider one
special type of structured information, i.e., the manifold structure
of images. In fact, this structured information can be explored in
sparse representation through Laplacian regularization
\cite{ZGL03,ZBLW04,lu2011latent}. Since the textual BOW model has been used for the above sparse representation, we define Laplacian regularization with the visual BOW model. The distinct advantage of using Laplacian
regularization is that we can induce the extra structured sparsity
into sparse representation and thus suppress the noise in the data.

To define the Laplacian regularization term for structured sparse
representation with respect to each image $x_i$, we first compute
the normalized Laplacian matrix as follows:
\begin{eqnarray}
L_i = I-D_i^{-1/2} Y_iY_i^TD_i^{-1/2},
\end{eqnarray}
where $Y_i=[Y_{j.}]_{j\in \mathcal{N}_k(i)}$ with $Y_{j.}$ being the
$j$-th row of the visual BOW model $Y$, and $D_i$ is a diagonal
matrix with its $j$-th diagonal element being the sum of the $j$-th
row of $Y_iY_i^T$. Here, we define the similarity matrix (i.e.
$Y_iY_i^T$) in the $k$-nearest neighborhood $\mathcal{N}_k(i)$ of
$x_i$ only with linear kernel. We further define the Laplacian
regularization term for the sparse representation problem in Equation (\ref{eq:sr}) as $\alpha_i^TL_i \alpha_i$. Let $V_i$ be a $k\times
k$ orthonormal matrix with each column being an eigenvector of
$L_i$, and $\Sigma_i$ be a $k \times k$ diagonal matrix with its
diagonal element $\Sigma_i(j,j)$ being an eigenvalue of $L_i$
(sorted as $\Sigma_i(1,1) \leq ...\leq \Sigma_i(k,k)$). Given that
$L_i$ is nonnegative definite, we have $\Sigma_i \geq 0$. Meanwhile,
since $L_i V_i = V_i \Sigma_i$ and $V_i$ is orthonormal, we have
$L_i = V_i \Sigma_i V_i^T$. Hence, $\alpha_i^TL_i \alpha_i$ can be reformulated as:
\begin{eqnarray}
\alpha_i^TL_i \alpha_i = \alpha_i^T V_i \Sigma_i^{\frac{1}{2}}
\Sigma_i^{\frac{1}{2}} V_i^T \alpha_i = \alpha_i^T \tilde{C}_i^T
\tilde{C}_i \alpha_i = ||\tilde{C}_i \alpha_i||_2^2,
\end{eqnarray}
where $\tilde{C}_i=\Sigma_i^{\frac{1}{2}}V_i^T$. That is, we have
successfully formulated $\alpha_i^TL_i \alpha_i$ as an $L_2$-norm
term.

However, if this Laplacian regularization term is directly added into the sparse representation problem in Equation~(\ref{eq:sr}), we would have difficulty in solving this problem efficiently. Hence, we further formulate an $L_1$-norm version of Laplacian regularization as:
\begin{eqnarray}
||\tilde{C}_i  \alpha_i||_1 =
||\Sigma_i^{\frac{1}{2}}V_i^T\alpha_i||_1.
\end{eqnarray}
By introducing noise terms for both linear reconstruction and
$L_1$-norm Laplacian regularization, we transform the sparse
representation problem in Equation~(\ref{eq:sr}) into
\begin{eqnarray}
&&\min_{\alpha_i,\zeta_i,\xi_i}~~||[\alpha_i^T,\zeta_i^T,
\xi_i^T]||_1,~~\nonumber \\
&&\mathrm{s.t.}~~y_i = C_i\alpha_i+\zeta_i,~0=\tilde{C}_i
\alpha_i+\xi_i, \label{eq:ssr}
\end{eqnarray}
where the reconstruction error and Laplacian regularization with
respect to $\alpha_i$ are controlled by $\zeta_i$ and $\xi_i$,
respectively. Let $\alpha'_i=[\alpha_i^T,\zeta_i^T, \xi_i^T]^T$,
$C'_i=\left[
  \begin{array}{ccc}
    C_i & I & 0 \\
    \tilde{C}_i & 0 & I \\
  \end{array}
\right]$, and $y'_i=[y_i^T,0^T]^T$. We finally solve the following
structured sparse representation problem for $L_1$-graph
construction:
\begin{eqnarray}
\min_{\alpha'_i}~~||\alpha'_i||_1,~~\mathrm{s.t.}~~y'_i = C'_i
\alpha'_i, \label{eq:ssrnew}
\end{eqnarray}
which takes the same form as the original sparse representation
problem in Equation~(\ref{eq:srsol}). The weight matrix $W$ of the $L_1$-graph $\mathcal{G} = \{\mathcal{V}, W\}$ can be defined the same as Equation (\ref{eq:l1wt}).

As we have mentioned, our $L_1$-norm Laplacian regularization can be
smoothly incorporated into the original sparse representation
problem in Equation (\ref{eq:sr}). However, this is not true for the traditional Laplacian regularization, which may introduce extra parameters (hard to tune in practice) into the $L_1$-optimization for sparse representation. More importantly, our $L_1$-norm Laplacian
regularization can induce the extra structured sparsity (i.e. the
sparsity of the noise term $\xi_i$), which is not ensured by the
traditional Laplacian regularization. It should be noted that the
$p$-Laplacian regularization \cite{ZS05} can also be regarded as an
$L_1$-generalization of Laplacian regularization with $p=1$. By
defining a matrix $C_i^{(p)} \in R^{\frac{k(k-1)}{2}\times k}$, the
$p$-Laplacian regularization can be formulated as $||C_i^{(p)}
\alpha_i||_1$ \cite{CLK10}, similar to our $L_1$-norm Laplacian
regularization. Hence, we can apply $p$-Laplacian regularization
similarly to structured sparse representation. However, this
Laplacian regularization causes much more time since
$C_i^{(p)}$ has a much larger size as compared to the matrix
$\tilde{C}_i \in R^{k\times k}$ used by our $L_1$-norm Laplacian
regularization proposed above.

\subsection{Efficient visual BOW refinement algorithm}
\label{sect:refine:alg}

After we have constructed the $L_1$-graph over images with
structured sparse representation, we further solve the visual BOW
refinement problem in Equation (\ref{eq:bowsr}) using the algorithm proposed in \cite{ZBLW04}. The complete algorithm for visual BOW refinement is outlined as follows:
\begin{description}
\item[(1)]
Construct an $L_1$-graph $\mathcal{G} = \{\mathcal{V}, W\}$ by
solving the structured sparse representation problem in Equation (\ref{eq:ssrnew}) in the $k$-nearest neighborhood of each image.
\item[(2)]
Compute the matrix $S=D^{-1/2}WD^{-1/2}$, where $D$ is a diagonal
matrix with its $i$-th diagonal element being the sum of the $i$-th
row of $W$.
\item[(3)]
Iterate $F(t+1) = \alpha S F(t)+(1-\alpha)Y$ for visual BOW
refinement until convergence, where the parameter
$\alpha=\lambda/(1+\lambda)$ (see the explanation below).
\item[(4)]
Output the limit $F^*$ of the sequence $\{F(t)\}$ as the final
refined visual BOW model.
\end{description}
According to \cite{ZBLW04}, the above algorithm converges to
$F^*=(1-\alpha)(I-\alpha S)^{-1}Y$, which is equal to Equation (\ref{eq:bowsrsol}) with $\alpha=\lambda/(1+\lambda)$. Since the structured sparse representation problem in Equation (\ref{eq:ssrnew}) is limited to $k$-nearest neighborhood, our $L_1$-graph construction in Step 1 has the same time complexity (with respect to the data size $n$) as $k$-NN graph construction. Moreover, given that $S$ is very sparse, Step 3 has a linear time complexity. Hence, the proposed algorithm can be applied to large datasets.

\begin{figure*}[t]
\vspace{0.00in}
\begin{center}
\includegraphics[width=0.76\textwidth]{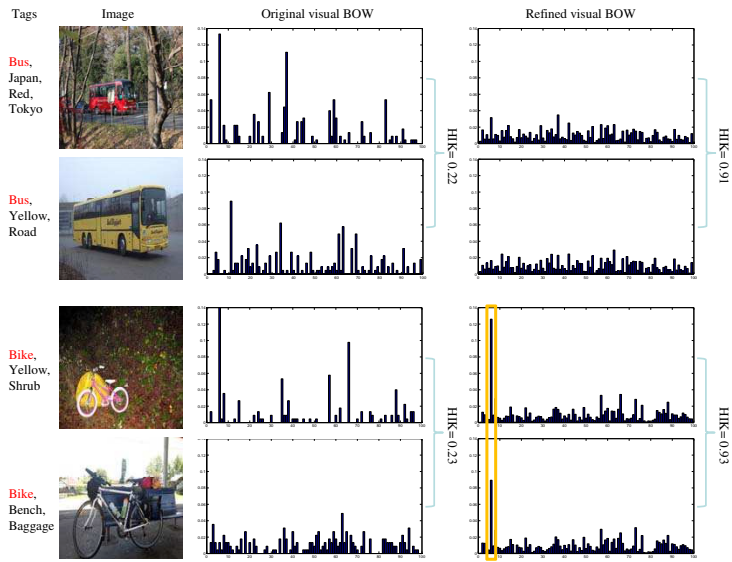}
\end{center}
\vspace{-0.25in} \caption{Illustrative comparison between the refined
and original visual BOW models. The similarity of two histograms is
measured by histogram-intersection kernel (HIK). It can be observed that the intra-class variations due to scale changes and cluttered backgrounds can be reduced by our visual BOW refinement. } \label{Fig.3} \vspace{-0.0in}
\end{figure*}

To give an explicit explanation of the refined visual BOW model
$F^*$, we show the illustrative comparison between the refined and
original visual BOW models in Figure~\ref{Fig.3}. Here, we conduct the
experiment of visual BOW refinement on a subset of the PASCAL VOC'07
dataset \cite{EVW07}. The immediate observation from Figure~\ref{Fig.3} is that the intra-class variations due to scale changes and cluttered backgrounds can be reduced by our visual BOW refinement in terms of HIK values. That is, the semantic gap associated with the original visual BOW is indeed bridged to some extent by exploiting the tags of images. Another interesting observation from Figure~\ref{Fig.3} is that some visual words of the refined visual BOW model tend to be explicitly related to the high-level semantics of images. For example, the visual word marked by an orange box is clearly shown to be related to ``bike", considering that this visual word becomes dominative (\emph{originally far from dominative}) in the histogram of the fourth image after visual BOW refinement. Such observation further verifies the effectiveness of
our visual BOW refinement.

\section{Visual BOW reduction}
\label{sect:reduct}

This section presents our visual BOW reduction on the refined visual
BOW model $F^*$ in detail. We first formulate visual BOW reduction
as a semantic spectral clustering problem, and then construct two
semantic graphs with the refined visual BOW model for such spectral
clustering. Finally, we provide the complete algorithm for visual
BOW reduction with the constructed semantic graphs.

\subsection{Problem formulation}

Since the semantic information has been successfully exploited for
visual BOW refinement in Section~\ref{sect:refine}, the semantic gap
associated with the visual BOW model can be bridged to some extent.
However, the vocabulary size of the refined visual BOW model remains
unchanged, which means that the subsequent image classification may
still take sheer amount of time given a large initial visual
vocabulary. In this paper, for efficient image classification, we
further reduce the refined visual BOW model to a much smaller size.
To explicitly preserve the manifold structure of mid-level features,
we formulate visual BOW reduction as spectral clustering over
mid-level features just as our short conference version \cite{LP11},
which is also shown in Figure~\ref{Fig.1}. The goal of spectral
clustering is to extract a reduced set of higher level features from
the original large vocabulary of mid-level features. In this paper,
spectral clustering of mid-level features is selected for visual BOW
reduction, because it can give explicit explanation of each reduced
feature (also see Figure~\ref{Fig.4}). However, this is not true for
the traditional dimension reduction methods \emph{directly using
spectral embedding over images} (or other similar techniques), since
the meaning of each dimension in the reduced space is unknown. In
the following, we will elaborate the key step of spectral clustering
used for our visual BOW reduction, i.e., \emph{spectral embedding
over mid-level features}.

Given a vocabulary of mid-level features $\mathcal{V}_m =
\{m_i\}_{i=1}^{M_v}$, we construct an undirected weighted graph
$\mathcal{G}_s=\{V_s,W_s\}$ with its vertex set $V_s =
\mathcal{V}_m$ and $W_s=[w^{(s)}_{ij}]_{M_v \times M_v}$, where
$w^{(s)}_{ij}$ denotes the similarity between two mid-level features
$m_i$ and $m_j$. In this paper, the weight matrix $W_s$ is computed
based on the refined visual BOW model $F^*$. Since spectral
embedding aims to represent each vertex in the graph as a lower
dimensional vector that preserves the similarities between the
vertex pairs, it is actually equivalent to finding the leading
eigenvectors of the normalized graph Laplacian
$L_s=I-D_s^{-1/2}W_sD_s^{-1/2}$, where $D_s$ is a diagonal matrix
with its $(i,i)$-element equal to the sum of the $i$-th row of
$W_s$. In this paper, we only consider this type of normalized
Laplacian \cite{NJW02}, regardless of other normalized versions
(e.g. \cite{LL06}). Let $\{(\lambda_i^{(s)}, \mathbf{v}_i^{(s)}):
i=1,...,M_v\}$ be the set of eigenvalues and the associated
eigenvectors of $L_s$, where $0 \leq \lambda_1^{(s)} \leq ... \leq
\lambda_{M_v}^{(s)}$ and $||\mathbf{v}_i^{(s)}||_2=1$. The
\emph{spectral embedding} of the graph $\mathcal{G}_s$ is
represented by
\begin{eqnarray}
E^{(K)}=(\mathbf{v}_1^{(s)},...,\mathbf{v}_{K}^{(s)}),
\end{eqnarray}
with the $j$-th row $E^{(K)}_{j.}$ being the new representation for
mid-level feature $m_j$. Given that $K<M_v$, the mid-level features
have actually been represented as lower dimensional vectors.

Since we have only formulated the key step of spectral clustering
(i.e. spectral embedding) in detail, we will further elaborate
\emph{graph construction for spectral clustering} and \emph{the
complete visual BOW reduction algorithm by spectral clustering} in
the next two subsections, respectively.

\subsection{Graph construction with refined visual BOW model}

In this subsection, we focus on graph construction for spectral
clustering of mid-level features. More concretely, based on the
refined visual BOW model $F^*$, we construct two graphs over
mid-level features, which are different only in how we quantify the
similarity between mid-level features. The details of the two graph
construction approaches are presented as follows.

The first graph $\mathcal{G}_s =\{\mathcal{V}_m,W_s\}$ is
constructed by defining the weight matrix $W_s=[w^{(s)}_{ij}]_{M_v
\times M_v}$ based on the Pearson product moment (PPM) correlation
\cite{RN88}. That is, given the refined visual BOW model $F^*$, the
similarity between mid-level features $m_i$ and $m_j$ is computed by
\begin{eqnarray}
w^{(s)}_{ij}=\frac{\sum_{n=1}^n
(F^*_{.i}-\mu_i)(F^*_{.j}-\mu_j)}{(n-1)\sigma_i\sigma_j},
\label{eq:swt1}
\end{eqnarray}
where $\mu_i$ and $\sigma_i$ are the mean and standard deviation of
$F^*_{.i}$ (the $i$-th column of $F^*$), respectively. If $m_i$ and
$m_j$ are not positively correlated, $w^{(s)}_{ij}$ will be
negative. In this case, we set $w^{(s)}_{ij}=0$ to ensure that the
weight matrix $W_s$ is nonnegative. Moreover, we construct the
second graph $\mathcal{G}_s =\{\mathcal{V}_m,W_s\}$ by directly
defining the weight matrix $W_s$ in the following matrix form:
\begin{eqnarray}
W_s=(F^*)^TAF^*, \label{eq:swt2}
\end{eqnarray}
where $A \in R^{n\times n}$ denotes the similarity (affinity) matrix
computed over the textual BOW model (also used as the weight matrix
in Equation (\ref{eq:lap})). Since the refined visual BOW model
$F^*$ used to construct the above two graphs has taken the tags of
images into account, we call them as semantic graphs. It should noted that these two graphs have a significant difference: the first graph is constructed only with the refined visual BOW model $F^*$, while the second graph exploits the textual BOW model besides $F^*$. More notable, the first method using Equation (\ref{eq:swt1}) for graph construction is proposed in our short conference paper \cite{LP11}, while the second method using Equation (\ref{eq:swt2}) is newly proposed in the present work. Since the textual BOW model is not used in Equation (\ref{eq:swt1}), the first method can handle images even without user-shared tags (just as \cite{LP11}), which is not the case for the second method using Equation (\ref{eq:swt2}). Meanwhile, our later experiments show that the second method for graph construction generally outperforms the first method due to the extra use of the textual BOW model.

The distinct advantage of using the above two graphs for spectral
clustering is that we have eliminated the need to tune any parameter
for graph construction which can significantly affect the
performance and has been noted as an inherent weakness of
graph-based methods. In contrast, the graph over mid-level features
is defined by a Gaussian function in \cite{LYS09} when each
mid-level feature is represented as a vector of point-wise mutual
information. As reported in \cite{LYS09}, the choice of the variance
in the Gaussian function affects the performance significantly. More
importantly, as shown in later experiments, our spectral clustering
with semantic graphs can help to discover more intrinsic manifold
structure of mid-level features and thus lead to obvious performance
improvements over \cite{LYS09}.

\subsection{Visual BOW reduction by semantic spectral clustering}
\label{sect:cse:cluster}

\begin{figure}[t]
\vspace{0.00in}
\begin{center}
\includegraphics[width=0.76\textwidth]{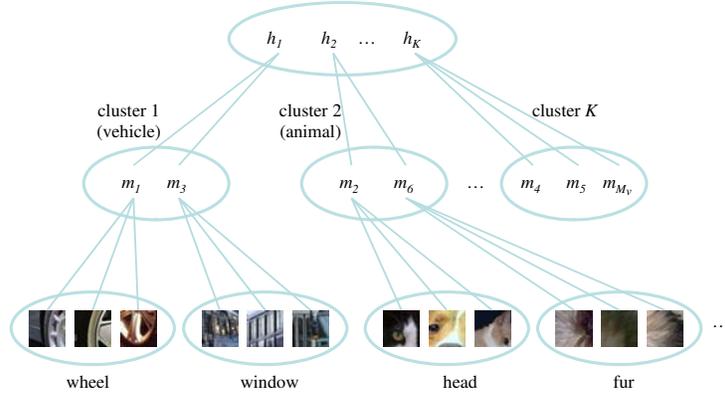}
\end{center}
\vspace{-0.25in} \caption{Illustration of the spectral clustering
results by our SSC for visual BOW reduction. The
reduced set of high-level features $\mathcal{V}_h=\{h_i\}_{i=1}^K$
are derived from the original visual vocabulary $\mathcal{V}_m =
\{m_j\}_{j=1}^{M_v}$, where each high-level feature refers
to a cluster of mid-level features. It can be observed that the
semantically related mid-level features tend to be grouped together
(e.g. wheel and window, or head and fur) by spectral clustering. }
\label{Fig.4} \vspace{-0.00in}
\end{figure}

After a semantic graph has been constructed based on the refined
visual BOW model, we perform spectral embedding on this graph. In
the new low-dimensional embedding space defined by Equation (13), we
reduce the visual vocabulary $\mathcal{V}_m$ into a set of higher
level features by $k$-means clustering. The complete algorithm for
visual BOW reduction is summarized as follows:
\begin{description}
\item[(1)]
Find $K$ smallest nontrivial eigenvectors $\mathbf{v}^{(s)}_1, ..., \mathbf{v}^{(s)}_K$ of $L_s=I-D_s^{-1/2}W_sD_s^{-1/2}$. Here, $W_s$ can be defined by either Equation (\ref{eq:swt1}) or Equation
(\ref{eq:swt2}).
\item[(2)]
Form $E^{(K)}=[\mathbf{v}^{(s)}_1,...,\mathbf{v}^{(s)}_K]$, and
normalize each row of $E^{(K)}$ to have unit length. Here, the
$i$-th row $E^{(K)}_{i.}$ denotes the new low-dimensional feature
vector for mid-level feature $m_i$.
\item[(3)]
Perform $k$-means clustering on the new low-dimensional feature
vectors $E^{(K)}_{i.}(i=1,...,M_v)$ to partition the vocabulary
$\mathcal{V}_m$ of $M_v$ mid-level features into $K$ clusters. Here,
each cluster of mid-level features denotes a new higher level
feature.
\end{description}
The above semantic spectral clustering algorithm is denoted as SSC
in the following. In particular, this algorithm is called SSC1 (or
SSC2) when the weight matrix is defined by Equation (\ref{eq:swt1})
(or Equation (\ref{eq:swt2})). It should be noted that our SSC
algorithm can run very efficiently even on a large dataset when the
data size $n\gg M_v$, since it has a time complexity of
$O({M_v}^3+K{M_v})$.

Let $\mathcal{V}_h=\{h_i\}_{i=1}^K$ be the reduced set of high-level
features which are learnt from the large vocabulary of mid-level
features by our SSC algorithm. According to the spectral clustering
results illustrated in Figure~\ref{Fig.4}, the relationships between
high-level and mid-level features can be represented using a single
matrix $U=[u_{ij}]_{K\times M_v}$, where $u_{ij}=1$ if mid-level
feature $m_j$ occurs in cluster $i$ (i.e. high-level feature $h_i$)
and $u_{ij}=0$ otherwise. The reduced visual BOW model $Y^*$ defined
over $\mathcal{V}_h$ can be readily derived from the refined visual
BOW model $F^*$ defined over $\mathcal{V}_m$ as follows:
\begin{eqnarray}
Y^*=F^*U^T.
\end{eqnarray}
As compared to the original visual BOW model $Y$, this reduced
visual BOW model $Y^*$ has two distinct advantages: the tags of
images have been added to it to bridge the semantic gap, and it has
a much smaller vocabulary size. Moreover, the semantic information
associated with the high-level features is also explicitly
illustrated in Figure~\ref{Fig.4}. We find that the semantically
related mid-level features are grouped together (e.g. wheel and
window, or head and fur) by spectral clustering and thus the
high-level features tend to be related to the semantics of images
(e.g. vehicle, or animal). In the following, the reduced visual BOW
model will be directly applied to image classification.

\section{Experimental results}
\label{sect:exp}

In this section, the proposed methods for visual BOW refinement and
reduction are evaluated in image classification. We first describe the experimental setup, including information of the two benchmark datasets and the implementation details. Moreover, our methods are compared with other closely related methods on the two benchmark datasets.

\subsection{Experimental setup}

\begin{figure}[t]
\vspace{0.00in}
\begin{center}
\includegraphics[width=0.80\textwidth]{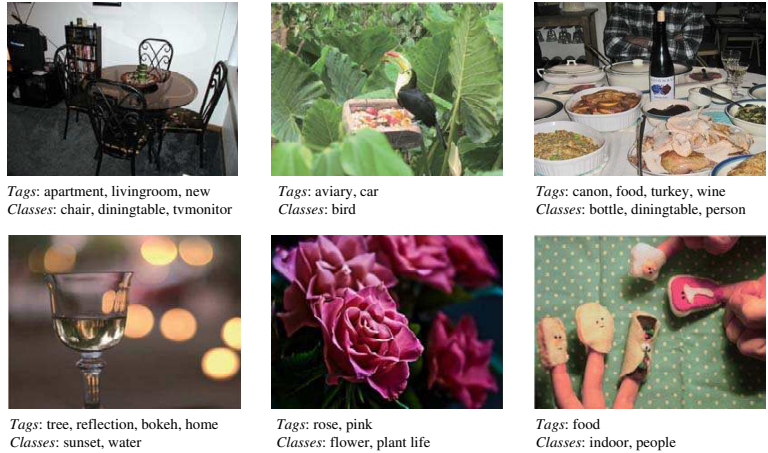}
\end{center}
\vspace{-0.25in} \caption{Example images from PASCAL VOC'07 (top row) and MIR FLICKR (bottom row) datasets with their tags and classes.}
\label{Fig.5} \vspace{-0.0in}
\end{figure}

We select two benchmark datasets for performance evaluation. The
first dataset is PASCAL VOC'07 \cite{EVW07} that contains around
10,000 images. Each image is annotated by users with a set of tags,
and the total number of tags used in this paper is reduced to 804 by
the same preprocessing step as \cite{GVS10}. This dataset is organized into 20 classes. Moreover, the second dataset is MIR FLICKR \cite{HL08} that contains 25,000 images annotated with 457 tags. This dataset is organized into 38 classes. For the PASCAL VOC'07 dataset, we use the standard training/test split, while for the MIR FLICKR dataset we split it into a training set of 12,500 images and a test set of the same size just as \cite{GVS10}. It should be noted that image classification on these two benchmark datasets is rather challenging, considering that each image may belong to multiple classes and each class may have large intra-class variations. Some example images from these two datasets are shown in Figure~\ref{Fig.5}.

For each dataset, we extract the same feature set as \cite{GVS10}. That is, we use local SIFT features \cite{Lowe04} and local hue histograms \cite{VWS06}, both computed on a dense regular grid and on regions found with a Harris interest-point detector. We quantize the four types of local descriptors using $k$-means clustering, and represent each image using four visual word histograms. Here, we only consider the four types of local descriptors to make a fair comparison with \cite{GVS10}, and other low-level visual features can be used similarly. Moreover, following the idea of \cite{LSP06}, each visual BOW representation is also computed over a $3 \times 1$ horizontal decomposition of the image, and concatenated to form a new representation that encodes some of the spatial layout of the
image. Finally, by concatenating all the visual BOW representations
into a single representation, we generate a large visual vocabulary
of about 10,000 mid-level features just as \cite{GVS10}. In our experiments, we only adopt $k$-means clustering for visual BOW generation, regardless of other clustering techniques \cite{Bagirov08,MNJ08}. Our main considerations are as follows: (1) we focus on visual BOW refinement and reduction, but not visual BOW generation; (2) we can make a fair comparison with the state-of-the-art, since $k$-means clustering is commonly used in related work (e.g. \cite{GVS10}).

\begin{figure*}[t]
\centering
\begin{minipage}{0.42\textwidth}
\centering
\includegraphics[width=0.96\textwidth]{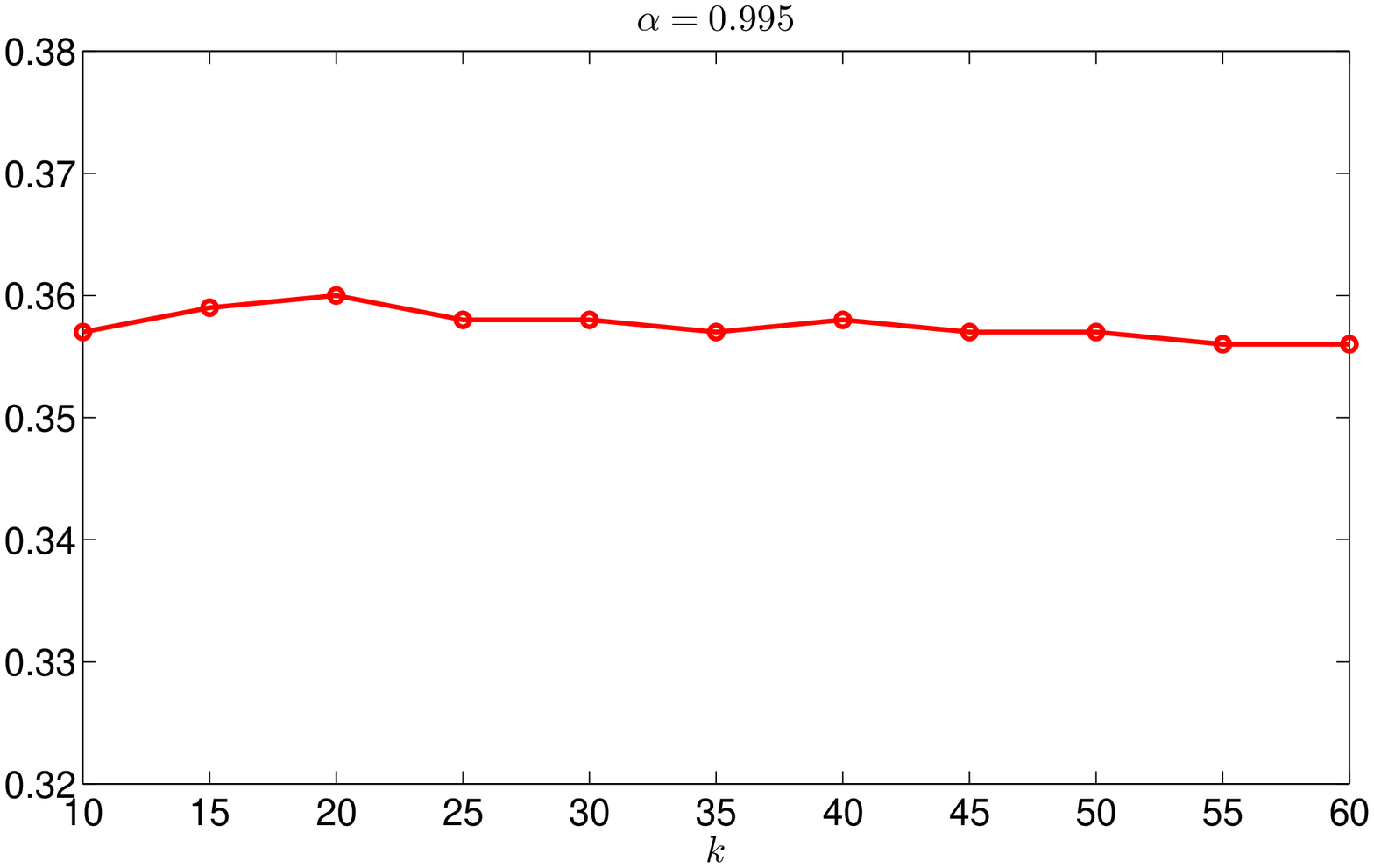}
\end{minipage}
\hspace{0.1in}
\begin{minipage}{0.42\textwidth}
\centering
\includegraphics[width=0.96\textwidth]{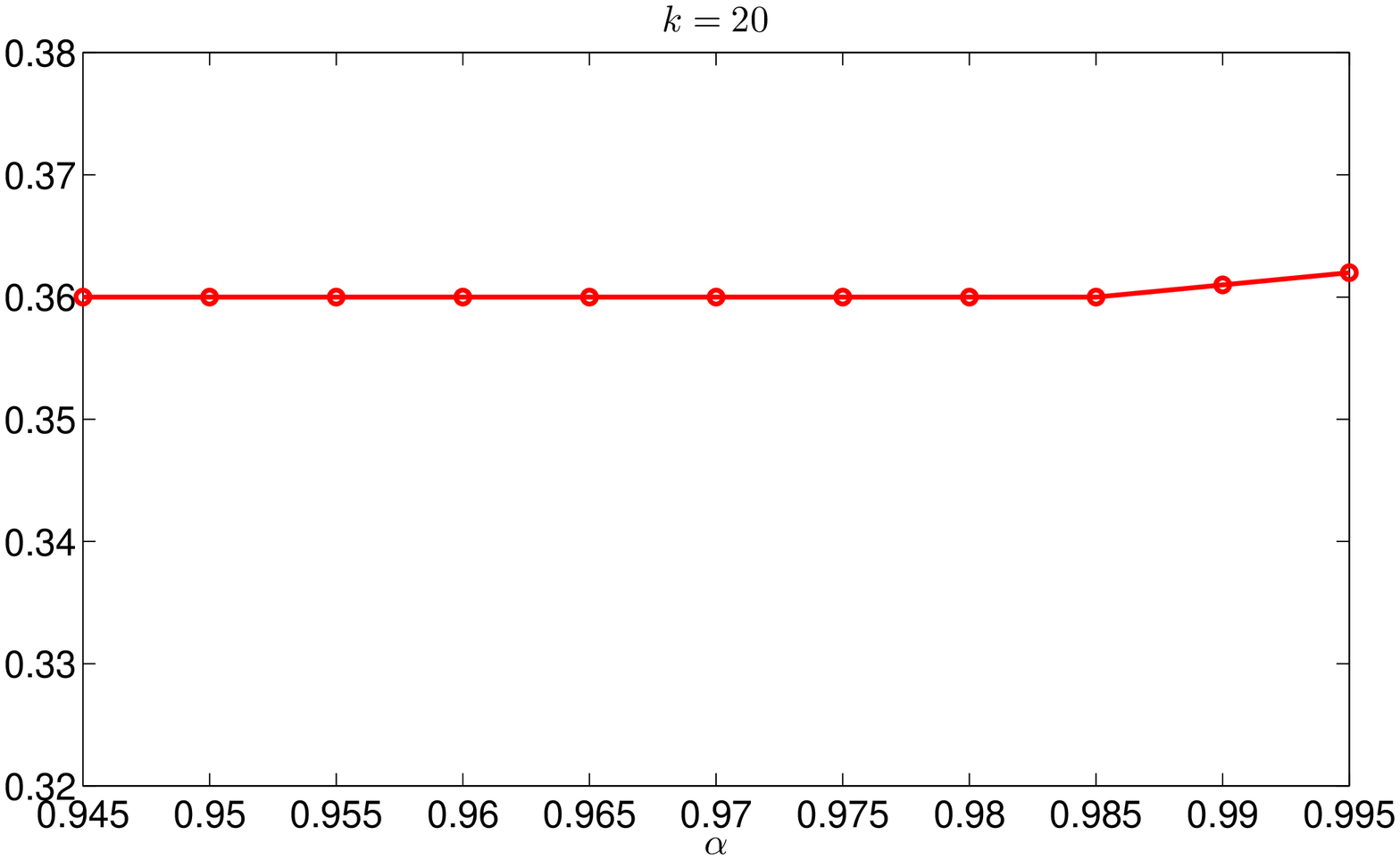}
\end{minipage}
\caption{The cross-validation classification results using the refined visual BOW models obtained by our SSL-SSR algorithm on the training set of the PASCAL VOC'07 dataset. } \label{Fig.6}
\vspace{-0.0in}
\end{figure*}

To evaluate the refined and reduced visual BOW models, we apply them
directly to image classification using SVM with $\chi^2$ kernel.
Since we actually perform multi-label classification on the two
benchmark datasets, the classification results are measured by mean
average precision (MAP) just the same as \cite{GVS10}. To show the
effectiveness of the refined visual BOW model obtained by our
graph-based SSL with structured sparse representation (SSL-SSR), we
compare it with the original visual and textual BOW models.
Moreover, our SSL-SSR is compared to graph-based SSL with sparse
representation (SSL-SR) and $k$-NN graph-based SSL (SSL-kNN). In the
experiments, the parameters of our SSL-SSR are selected by
cross-validation on the training set. For example, according to
Figure~\ref{Fig.6}, we set the two parameters of our SSL-SSR on the
PASCAL VOC'07 dataset as: $k=20$ and $\alpha=0.995$ (which appear in
Steps 1 and 3 of our algorithm proposed in Section \ref{sect:refine:alg}). In fact, our algorithm is shown to be not much sensitive to the choice of these parameters, and we select relatively smaller values for $k$ to ensure its efficient running. For fair comparison, the same parameter selection strategy is used for other visual BOW refinement methods. Finally, our two SSC methods for visual BOW reduction are compared with diffusion map (DM) \cite{LYS09}, locally linear embedding (LLE), Eigenmap, latent Dirichlet allocation (LDA) \cite{BNJ03}, and principal component analysis (PCA). Here, SSC, DM, LLE, and Eigenmap are used for visual BOW reduction based on nonlinear manifold learning, while LDA actually ignores the manifold structure of mid-level features and PCA is only a linear dimension reduction method. For fair comparison, these methods for visual BOW reduction are all performed over the refined visual BOW model obtained by our refinement method SSL-SSR.

\begin{figure*}[t]
\vspace{0.00in}
\begin{center}
\subfigure[]{
\includegraphics[width=0.42\textwidth]{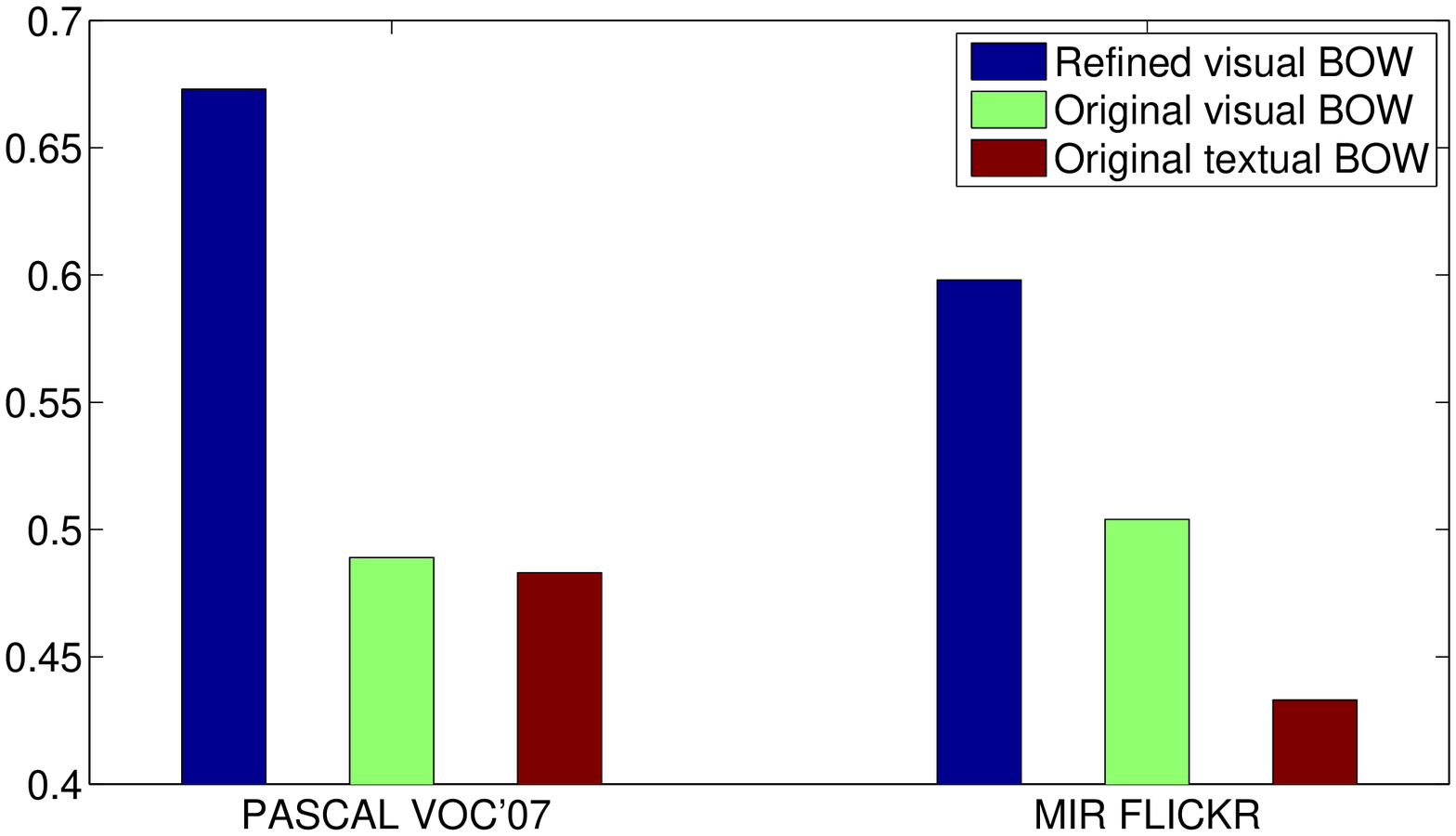}
\label{Fig.7:a}} \hspace{0.15in} \subfigure[]{
\includegraphics[width=0.42\textwidth]{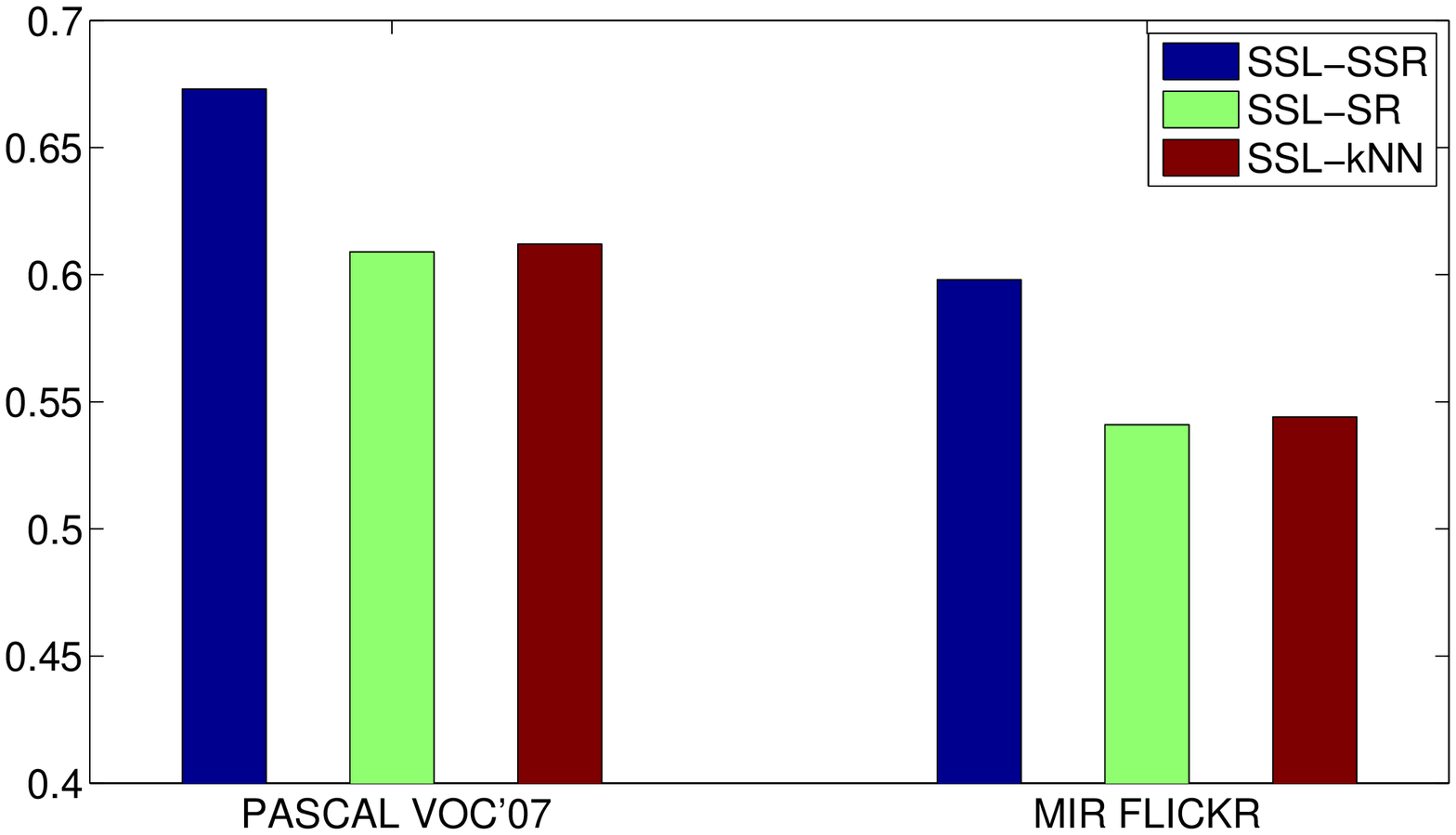}
\label{Fig.7:b}}
\end{center}
\vspace{-0.3in} \caption{The test classification results using the
refined visual BOW models on the two benchmark datasets: (a)
comparison between different BOW models; (b) comparison between
different visual BOW refinement methods.} \label{Fig.7}
\vspace{-0.0in}
\end{figure*}

\subsection{Results of visual BOW refinement}

To verify the effectiveness of our refined visual BOW model in image
classification, we show the comparison between different BOW models
in Figure~\ref{Fig.7:a}. The immediate observation is that the refined
visual BOW model by our SSL-SSR algorithm significantly outperforms
(38\% gain for PASCAL VOC'07 and 19\% gain for MIR FLICKR) the
original visual BOW model. That is, the tags of images have been
successfully added to the refined visual BOW model and thus the
semantic gap associated with the original visual BOW model has been
bridged effectively. More notably, our refined visual BOW model is
even shown to achieve more than 38\% gains over the original textual
BOW model for both of the two benchmark datasets. The significant
gains over both the original visual and textual BOW models are due
to the fact that our SSL-SSR algorithm can exploit these two types
of BOW models simultaneously for visual BOW refinement. In other
words, the original visual and textual BOW models can complement
each other well for image classification. This is also the reason
why we have made much effort to explore them not only in graph-based
SSL but also in $L_1$-graph construction.

The comparison between different visual BOW refinement methods is
further shown in Figure~\ref{Fig.7:b}. In this paper, to effectively
explore both visual and textual BOW models in graph construction, we
have developed a new $L_1$-graph construction method using
structured sparse representation (SSR), which is limited to
$k$-nearest neighborhood so that the $L_1$-graph can be constructed
efficiently even on large datasets. From Figure~\ref{Fig.7:b}, we find
that our SSR method can achieve about 10\% gains over the other two
graph construction methods (i.e. SR and kNN). These impressive gains
are due to the fact that the manifold structure of images derived
from the original visual BOW model can be explored by $L_1$-norm
Laplacian regularization to suppress the negative effect of noisy
tags, while such important structured information is completely
ignored by the other two graph construction methods.

\begin{figure*}[t]
\vspace{0.00in}
\begin{center}
\includegraphics[width=0.75\textwidth]{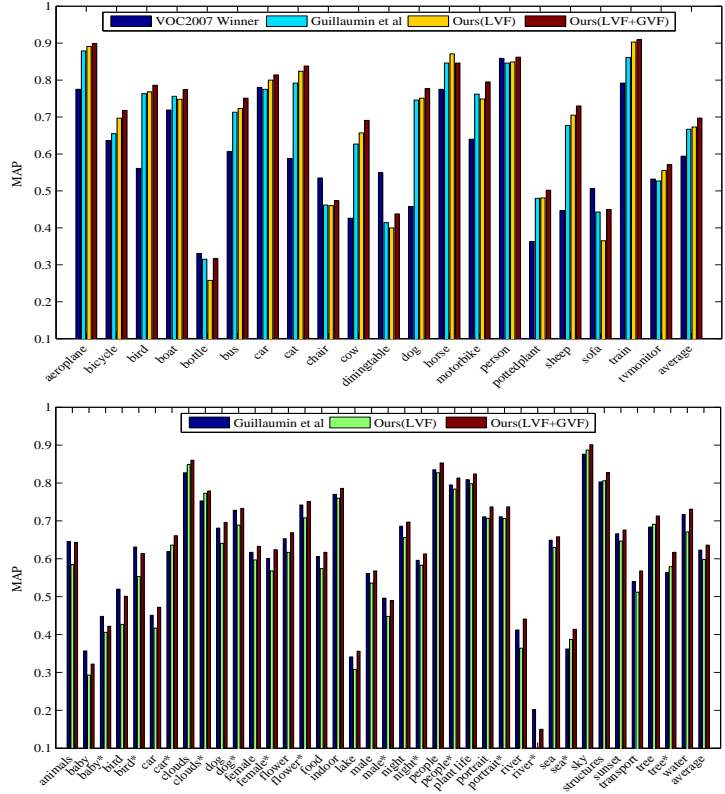}
\end{center}
\vspace{-0.3in}\caption{Comparison of our SSL-SSR method with the
state-of-the-art with respect to individual classes on PASCAL VOC'07 (top row) and MIR FLICKR (bottom row) datasets. Here, a class marked with * means it should be salient within images.} \label{Fig.8} \vspace{-0.0in}
\end{figure*}

\begin{table}[t]
\vspace{0.05in} \caption{Comparison of our SSL-SSR method with the
state-of-the-art on the two benchmark datasets (LVF: local visual
features; GVF: global visual features)} \label{Table.1}
\vspace{-0.0in}
\begin{center}
\tabcolsep0.1cm
\begin{tabular}{|c|ccc|cc|}
\hline
Methods      & LVF & GVF &  Tags  & PASCAL VOC'07 & MIR FLICKR \\
\hline
Winner       &  yes  &  yes  &  no   & 0.594  &  --   \\
\cite{GVS10} & yes & yes & yes & 0.667  & 0.623 \\
Ours (LVF)   &  yes  &  no  &  yes  & 0.673  &  0.598 \\
Ours (LVF+GVF)  & yes & yes & yes & \textbf{0.697} & \textbf{0.636}\\
\hline
\end{tabular}
\end{center}
\vspace{-0.0in}
\end{table}

The comparison of our SSL-SSR method with the state-of-the-art on
the two benchmark datasets is shown in Table~\ref{Table.1} and Figure~\ref{Fig.8}. To the best of our knowledge, the recent work \cite{GVS10} has reported the best results so far for image classification on the PASCAL VOC'07 and MIR FLICKR datasets. However, when the refined visual BOW model (i.e. local visual features) obtained by our method is fused with the global visual features (i.e. color histogram and GIST descriptor \cite{OT01}), our method is shown to achieve better results than \cite{GVS10} on both of the two benchmark datasets. By further observation over individual classes, we find that our method outperforms \cite{GVS10} on most classes. This becomes more impressive given that the present work makes use of much weaker global visual features than \cite{GVS10} (i.e. two types vs. seven types). Moreover, from Table~\ref{Table.1} and Figure~\ref{Fig.8}, we can also observe that both \cite{GVS10} and our method obviously outperform the VOC'07 winner due to the effective use of extra tags for image classification.

\begin{figure*}[t]
\vspace{0.00in}
\begin{center}
\subfigure[]{
\includegraphics[width=0.40\textwidth]{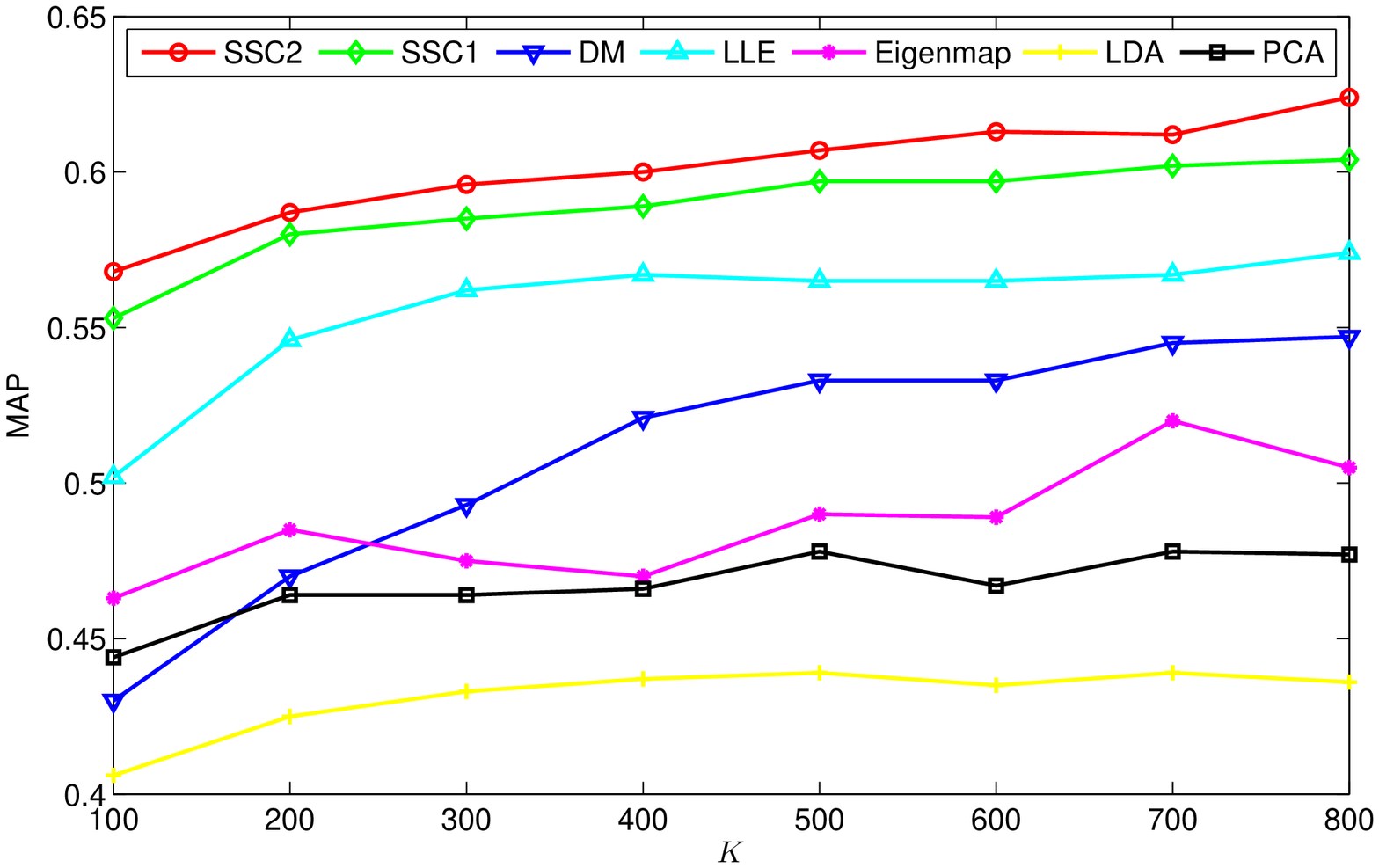}
\label{Fig.9:a}} \hspace{0.15in} \subfigure[]{
\includegraphics[width=0.40\textwidth]{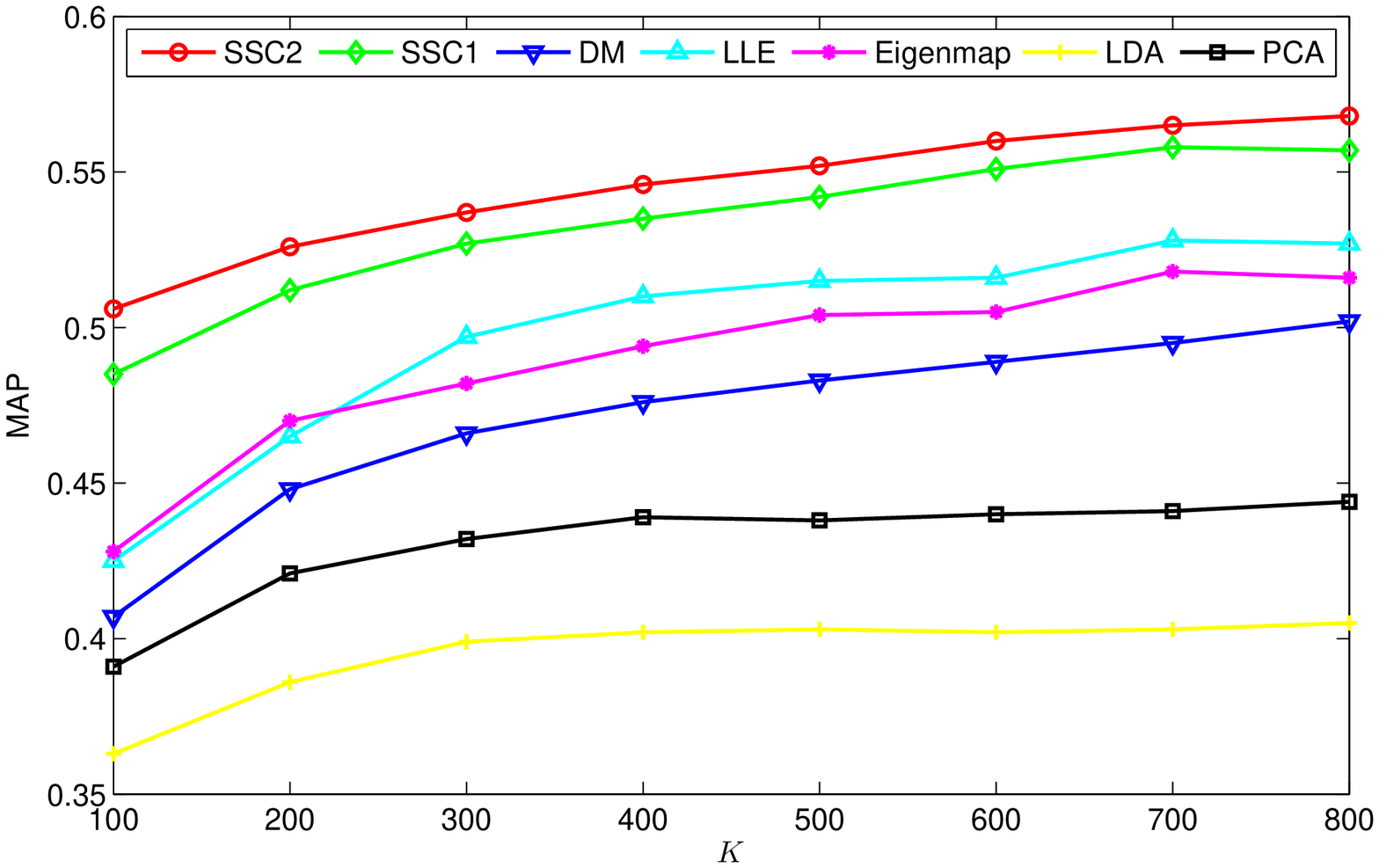}
\label{Fig.9:b}}
\end{center}
\vspace{-0.3in} \caption{Comparison between different visual BOW
reduction methods when a varied number of high-level features are
learnt from the refined visual BOW model of large vocabulary size:
(a) PASCAL VOC'07; (b) MIR FLICKR.} \label{Fig.9} \vspace{-0.05in}
\end{figure*}

\subsection{Results of visual BOW reduction}

In the above subsection, we have just verified the effectiveness of
our visual BOW refinement. In this subsection, we will further
demonstrate the promising performance of our visual BOW reduction.
The comparison between different visual BOW reduction methods is
shown in Figure~\ref{Fig.9}, where the five reduction methods are
performed over the same refined visual BOW model obtained by our
SSL-SSR algorithm. We first find that our SSC2 generally outperforms
our SSC1, due to the extra use of the textual BOW model in graph
construction for our SSC2. Moreover, we find that our SSC methods
always outperform the other three nonlinear manifold learning methods (i.e. DM, LLE, Eigenmap) for visual BOW reduction \cite{LYS09}. The reason is that our SSC methods have eliminated the need to tune any parameter for graph construction which can significantly affect the performance and has been noted as an inherent weakness of graph-based methods. Finally, as compared to the topic model LDA without considering the manifold structure of mid-level features and the linear dimension reduction method PCA, our SSC methods using nonlinear manifold learning for visual BOW reduction are shown to achieve significant gains in all cases. This is really very expressive, considering the outstanding performance of LDA and PCA in dimension reduction which has been extensively demonstrated in the literature.

\begin{figure*}[t]
\vspace{0.02in}
\begin{center}
\subfigure[]{
\includegraphics[width=0.42\textwidth]{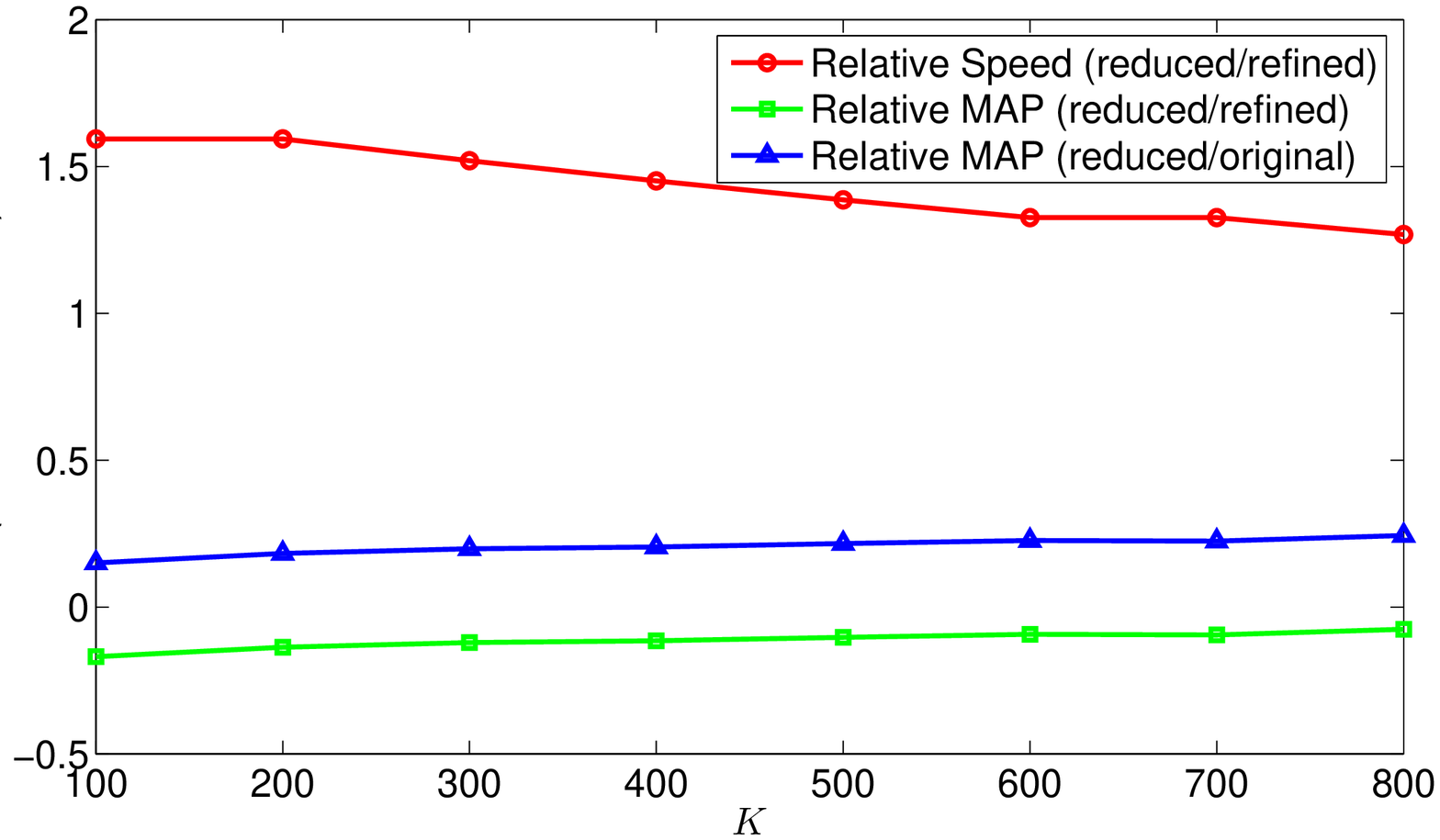}
\label{Fig.10:a}} \hspace{0.15in} \subfigure[]{
\includegraphics[width=0.42\textwidth]{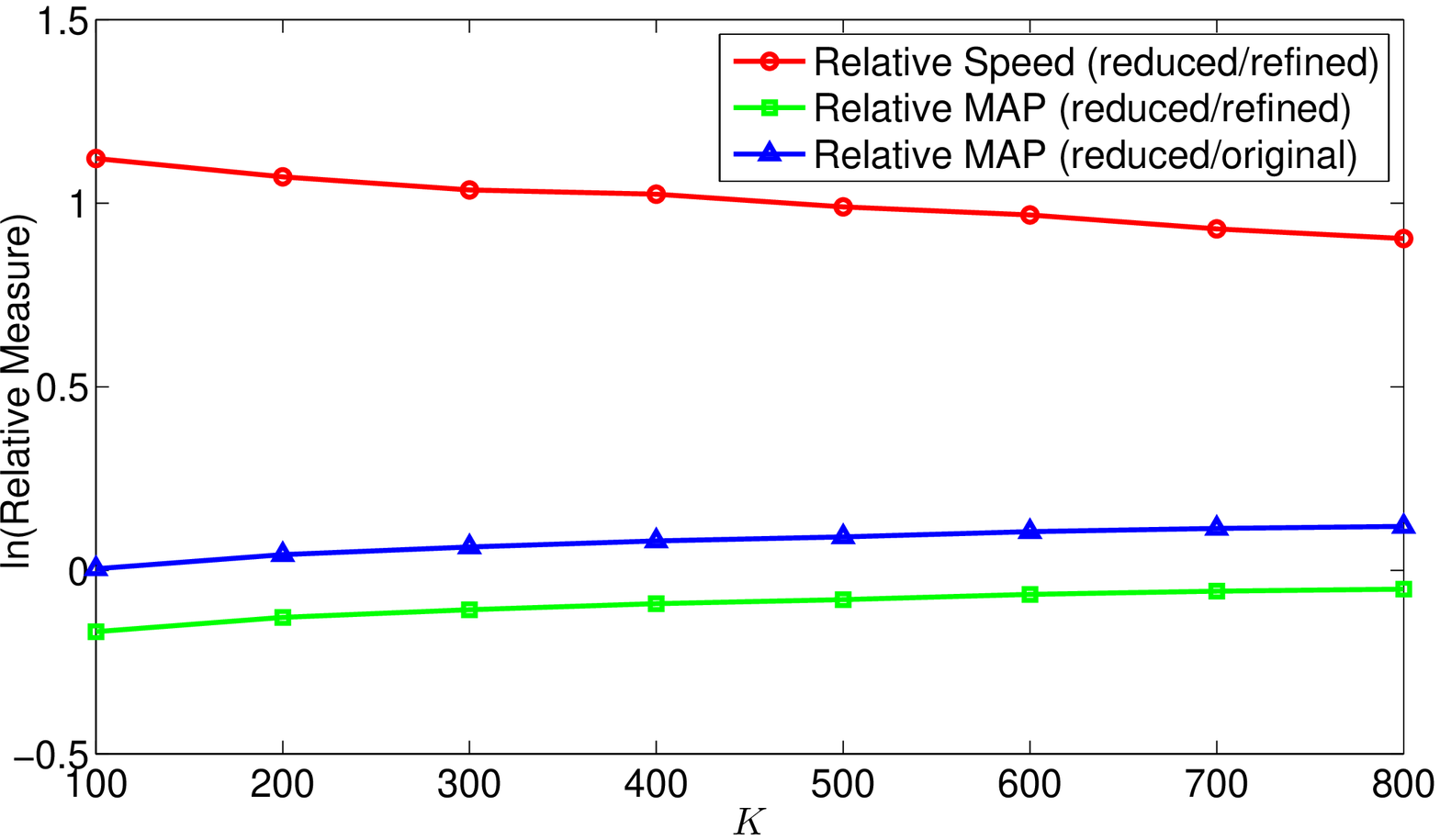}
\label{Fig.10:b}}
\end{center}
\vspace{-0.3in} \caption{Comparison of the reduced visual BOW model
obtained by our SSC2 method with the refined and original visual BOW
models in terms of different relative measures: (a) PASCAL VOC'07;
(b) MIR FLICKR. Here, the \emph{logarithmic plot} is used to show
different relative measures in a single figure for our SSC2 method.}
\label{Fig.10} \vspace{-0.0in}
\end{figure*}

Although we have just shown the significant gains achieved by our
SSC methods over other visual BOW reduction methods, we still need to compare the reduced visual BOW model obtained by
our SSC methods with the refined and original visual BOW models so
that we can directly verify the effectiveness of our reduced visual
BOW model in image classification. Figure~\ref{Fig.10} shows different
relative measures of our SSC2 method for visual BOW reduction. Here,
both relative speed and MAP measures are computed over the refined
and original visual BOW models, respectively. In particular, the
efficiency of visual BOW reduction is actually measured by the speed
of image classification with SVM. It should be noted that we only
show ``Relative Speed (reduced/refined)'' in Figure~\ref{Fig.10}
without considering ``Relative Speed (reduced/original)'', since the
refined and original visual BOW models cause comparable time in
image classification. From Figure~\ref{Fig.10}, we find that our method can reduce the visual vocabulary size (around 10,000) to a very low
level (e.g. $K=100$) and thus speed up the run of SVM significantly
(e.g. 400\% and 200\% faster for the two datasets, respectively),
but without decreasing MAP obviously as compared to the refined
visual BOW model. Moreover, our method is even shown to achieve
speed and MAP gains simultaneously over the original visual BOW
model. When the total time taken by our method (including visual BOW
refinement, visual BOW reduction, and image classification with the
reduced BOW model) is considered, we find from Table~\ref{Table.2}
that our method still runs more efficiently (e.g. 38\% faster) than
image classification directly with the original visual BOW model.
Here, we run the algorithms (Matlab code) on a computer with 3GHz CPU
and 32GB RAM. Finally, it is noteworthy that our method can play a
more important role in other image analysis tasks (e.g. image
retrieval) that have a higher demand of real-time response during
the testing or querying stage.

\begin{table}[t]
\vspace{0.05in} \caption{Comparison between the total time (minutes)
taken by the reduced ($K=100$) and original visual BOW models for
image classification with SVM on MIR FLICKR}
\label{Table.2} \vspace{-0.1in}
\begin{center}
\tabcolsep0.21cm
\begin{tabular}{|c|ccc|c|}
\hline
Visual BOW models    & Refinement &  Reduction  &  Classification  & Total \\
\hline
Original visual BOW &  --    &  --   &   47.9  &   47.9 \\
Reduced visual BOW  &  13.4  &  5.7  &   15.6  &   34.7 \\
\hline
\end{tabular}
\end{center}
\vspace{-0.2in}
\end{table}

\section{Conclusions}
\label{sect:con}

We have proposed a new framework for visual BOW refinement and
reduction to overcome the two drawbacks associated with the visual
BOW model. To overcome the first drawback, we have developed a
graph-based SSL method with structured sparse representation to
exploit the tags of images (easy to access for social
images) for visual BOW refinement. More importantly, for efficient
image classification, we have further developed a semantic spectral
clustering algorithm to reduce the refined visual BOW model to a
much smaller size. The effectiveness of our visual BOW refinement
and reduction has been verified by the extensive results on the PASCAL VOC'07 and MIR FLICKR benchmark datasets. In particular, when the global visual features are fused with visual BOW models, we can obtain the best results so far (to the best of our knowledge) on the two benchmark datasets.

The present work can be further improved in the following ways: (1) our visual BOW refinement and reduction can be readily extended to other challenging applications such as cross-media retrieval and social image parsing; (2) the depth information can be similarly used by our algorithms instead of the tags of images, which plays an important role in the ImageCLEF Robot Vision task; (3) our main ideas can be used reversely to refine the textual information using visual content.

\section*{Acknowledgments}

This work was supported by National Natural Science Foundation of
China under Grants 61202231 and 61222307, National Key Basic Research Program (973 Program) of China under Grant 2014CB340403, Beijing Natural Science Foundation of China under Grant 4132037, Ph.D. Programs Foundation of Ministry of Education of China under Grant 20120001120130, the Fundamental Research Funds for the Central Universities and the Research Funds of Renmin University of China under Grant 14XNLF04, and a grant from Microsoft Research Asia.

{\small
\bibliographystyle{elsarticle-num}

}

\end{document}